\documentclass[runningheads]{llncs}
\usepackage[T1]{fontenc}
\usepackage{multirow}
\usepackage{graphicx}
\usepackage{array}
\begin{document}
\renewcommand{\inst}[1]{\leavevmode\unskip$^{#1}$}
\title{Artificial Intelligence Bias on English Language Learners in Automatic Scoring}
\titlerunning{Artificial Intelligence Bias on English Language Learners in ...}
%
\author{Shuchen Guo\textsuperscript{*,\dag}\inst{1,2} \and
Yun Wang\textsuperscript{\dag}\inst{3} \and
Jichao Yu\textsuperscript{\dag}\inst{3} \and
Xuansheng Wu\textsuperscript{\dag}\inst{3} \and
Bilgehan Ayik\inst{4}\and
Field M. Watts\inst{5}\and
Ehsan Latif\inst{2}\and
Ninghao Liu\inst{3} \and
Lei Liu\inst{5} \and
Xiaoming Zhai\textsuperscript{*}\inst{2}}


\authorrunning{S. Guo et al.}
%

\institute{
\inst{1} School of Teacher Education, Nanjing Normal University, Nanjing Jiangsu 210023, China \\
\inst{2} AI4STEM Education Center, University of Georgia, Athens GA 30602, USA \\
\inst{3} School of Computing, University of Georgia, Athens, GA 30602, USA \\
\inst{4} College of Education and Human Development, George Mason University, Fairfax, VA 22030, USA \\
\inst{5} Educational Testing Service, Princeton, NJ, USA \\
\email{Corresponding to: xiaoming.zhai@uga.edu, 69010@njnu.edu.cn}
}

%
\maketitle
%
\renewcommand{\thefootnote}{\dag}
\footnotetext{These authors contributed equally to this work.\\
\textbf{Cite the study:} Guo, S., Wang, Y., Yu, J. Wu, X., Ayik, B., Watts, F., Latif, E., Liu, L., Liu, N., \& Zhai, X. (2025). Artificial Intelligence Bias on English Language Learners in Automatic Scoring. Proceedings of the International Conference on AI in Education, pp. 1-12, Palermo, Italy. }
\renewcommand{\thefootnote}{\arabic{footnote}}
\begin{abstract}
This study investigated potential scoring biases and disparities toward English Language Learners (ELLs) when using automatic scoring systems for middle school students' written responses to science assessments. We specifically focus on examining how unbalanced training data with ELLs contributes to scoring bias and disparities. We fine-tuned BERT with four datasets: responses from (1) ELLs, (2) non-ELLs, (3) a mixed dataset reflecting the real-world proportion of ELLs and non-ELLs (unbalanced), and (4) a balanced mixed dataset with equal representation of both groups. 
The study analyzed 21 assessment items: 10 items with about 30,000 ELL responses, five items with about 1,000 ELL responses, and six items with about 200 ELL responses. Scoring accuracy (Acc) was calculated and compared to identify bias using Friedman tests. We measured the Mean Score Gaps (MSGs) between ELLs and non-ELLs and then calculated the differences in MSGs generated through both the human and AI models to identify the scoring disparities.
We found that no AI bias and distorted disparities between ELLs and non-ELLs were found when the training dataset was large enough ($\textnormal{ELL}\approx30,000$ and $\textnormal{ELL}\approx1,000$), but concerns could exist if the sample size is limited ($\textnormal{ELL} \approx 200$).

\end{abstract}
\section{Introduction}

One of the most prominent innovations in assessment practices is the use of artificial intelligence (AI) for automatic scoring \cite{guo2024artificial}.  
AI could provide real-time scores and feedback to students, saving teachers' time and reducing their labor burden ~\cite{guo2024using,zhai2020applying}. 
While AI-based automatic assessment has brought significant potential to revolutionize assessment practices, concerns about scoring bias have also arisen~\cite{zhai2023ai}. Research~\cite{Latif2023,li2024using} has found that AI may yield deviated results in favor of students from certain groups, which is detrimental to equity and fairness. Among various possible factors that can result in scoring bias, such as gender \cite{Latif2023} and race, students' English proficiency has also gained special attention recently~\cite{wilson2024validity}. 

English language learners (ELLs) have unique strengths in learning due to their diverse backgrounds, and AI scoring systems should leverage this diversity. 
However, several researchers raised concerns about potential bias embedded in automated essay scoring systems while evaluating the ELLs' work \cite{weigle2013english,wilson2024validity}.
There are also empirical studies that found that AI scoring models are usually trained with natural language materials with the ability to score most student writings, but vary in the ability to score ELLs’ responses \cite{liang2023gpt}. 
Researchers have suggested potential bias can arise from the training samples, which often consist predominantly of non-ELLs, with much fewer ELLs represented \cite{wilson2024validity}. Compared to non-ELLs, ELLs exhibit unique linguistic features and writing styles, which can contribute to AI scoring biases, particularly when the models are trained with imbalanced datasets with a majority of responses from non-ELLs. \cite{leclair2009english,jourdain2016language}. 

While much research has been conducted on AI bias in language education, particularly in essay scoring, little is known about this issue in science assessments. Unlike essay writing tasks in language learning, student responses in science assessments are typically much shorter. The primary goal of these assessments is to evaluate students' science thinking rather than language proficiency ~\cite{zhai2024ai}. As such, it becomes crucial to examine whether and how AI scoring might introduce scoring bias for ELLs in the context of science assessments.

This study investigates whether the AI automatic scoring system exacerbates bias and disparities between ELLs and non-ELLs in science assessments. We fine-tuned a series of BERT models using training datasets that consisted of ELLs, non-ELLs, unbalanced mixed datasets, and balanced mixed datasets. These models were applied to 21 constructed response assessment items across various data scales. We then compared model accuracy and mean score gaps (MSGs) to analyze potential scoring bias and disparities. The study aimed to answer the following research question: How do training samples impact the AI scoring bias and disparities for ELLs?

\section{Related Work}
\subsection{Automatic Scoring in Science Assessment}
In response to the call for promoting three-dimensional (3D) science learning—integrating disciplinary core ideas, crosscutting concepts, and scientific practices, researchers advocate for developing assessment items that move beyond rote memorization tasks and require the application of knowledge and higher-order thinking skills. These assessment items are often performance-based constructed responses. Students are asked to write constructed responses to express their ideas \cite{zhai2020applying}. Compared to written tasks in language learning, the constructed responses have unique characteristics. On the one hand, student responses are typically short in science assessments, which provides limited information for AI scoring. On the other hand, students are often asked to explain phenomena or make an argument, etc. Based on students' responses, AI scoring aims to assess students' performance in these practices (e.g., explanation or argumentation). 

However, one challenge of using constructed response assessments in science teaching is the significant time commitment required for teachers to score student-constructed responses. Nowadays, the challenge is addressed by AI-powered automatic scoring.
Automated scoring can significantly reduce the time and effort required to score large numbers of student responses, which can be particularly useful in educational settings or standardized testing, where manual grading would be time-consuming and labor-intensive \cite{bablu2024machine,ifenthaler2022automated}. Additionally, in classroom settings, automated scoring plays a pivotal role in meeting the pressing need for efficient, precise, and timely assessment of students’ thinking, which serves as indispensable information provided to students and/or teachers to support adaptive learning \cite{lee2019automated,gerard2016using}. Studies have demonstrated the potential of AI-based automatic scoring to assess student written responses in science assessment using various AI technologies, including machine learning \cite{lee2019automated} and large language models \cite{lee2024applying} for both text-based \cite{lee2024applying,lee2019automated} and image-based responses \cite{zhai2022applying}. 
These studies show that AI scoring models can achieve a level of similarity to human scores. However, there is still ample opportunity for improvement \cite{zhai2020applying}. 

\subsection{AI Bias and Disparities in Automatic Scoring}

AI scoring bias is one of the major concerns researchers have when developing and using automatic scoring techniques \cite{wilson2024validity,li2024using}.
AI bias refers to systematic rather than random error. 
To detect AI bias, studies use the scoring accuracy (Acc) of the AI model as an indicator. 
Meanwhile, AI disparities can be defined as the difference in Mean Score Gaps (MSGs) between human and AI scores for responses written by different groups of students \cite{Latif2023}. When human MSGs and AI model MSGs are the same, it indicates no disparity. However, if the AI model MSG is significantly different from the human MSG, it means that using the AI scoring model can distort the scoring pattern for different groups of students  \cite{Latif2023}.

\subsection{AI Bias and Disparities for ELLs}
Bias and uneven treatment of ELLs and non-ELLs have always been a concern in traditional educational environments. For instance, science teachers in the US, who are mostly English speakers, can view ELLs through a deficit lens and thus expect less from them and assess them according to those lower expectations \cite{li2024using}. With the introduction of AI in science education, it is not surprising that educators and researchers have expressed concerns about whether the use of AI may exacerbate the existing bias in education that mostly impacts minoritized ELL groups \cite{cheuk2021can}.
Theoretically, the AI scoring model could have a statistical and computational bias if the training dataset lacks representation of certain groups, since the AI scoring model would reflect the pattern of the training dataset \cite{schwartz2022towards}. When it comes to ELLs, although the proportion of ELL students is increasing rapidly in some multicultural and multilingual countries like the US, the percentage of ELLs compared to non-ELLs is significantly much lower (e.g., 1:9 in the US), which would result in unbalanced datasets for AI model training. Meanwhile, the characteristics of linguistic features of ELLs and non-ELLs are obvious, which makes it a potential pattern to influence AI model scoring on ELLs' responses. AI may struggle to interpret nonstandard or unexpected responses of ELLs, resulting in misinterpretation or underestimation of valid yet diverse expressions of knowledge from students with different English proficiency levels \cite{cheuk2021can}. This situation can perpetuate existing inequities in educational assessment.

Many empirical studies have found indispensable AI bias for students of different English proficiency levels. For example, research \cite{jeon2024rethinking} stated that the speech recognition features of current chatbots are mostly sensitive to the pronunciation of native speakers and have trouble identifying other utterances, such as those from non-native speakers. A recent study from Stanford University demonstrates that there is a notable difference in false flags for non-native English speakers compared to native speakers and that AI-detectors are very prone to mistakenly identify non-native English speakers' writing as AI-generated, and thus considering them as a paradigm or cheating \cite{liang2023gpt}. The Center for Democracy and Technology (CDT) in the US has also released a report stating that ELLs are at particular risk of unequal treatment inside and outside the classroom due to the increasing use of generative AI. 
Particularly in science assessment, Li et al. \cite{li2024using} compared AI-generated scores using Convolutional Neural Networks (CNN) and human-generated scores of scientific models drawn by elementary school Multilingual Language Learners (MLLs) and found that AI-generated scores were more in the mid-range, while teachers provided scores more evenly spread across the scoring scale, which may reflect a more nuanced interpretation of the models.

It is reasonable to consider student English proficiency as an impact factor that may influence AI scoring performance, especially in science assessment, where the main goal is to assess student science. 
The different writing for ELLs could make it more challenging for AI models to accurately evaluate their actual performance in science. Therefore, empirical studies are needed to investigate whether and how an AI scoring model would have a bias for ELLs.

\section{Methods}
\subsection{Dataset}
The dataset used in this study is from the California Science Test (CAST), a state-administered assessment designed to measure students' knowledge and abilities in science. The CAST is administered to students in grade five, grade eight, and high school. For this study, we utilized 21 constructed response items from the Grade 8 test. In response to each item, students are required to write some words or sentences. Each item is scored on a scale of 0, 1, or 2, with scores typically based on the correctness of the response.


Within the data, students are categorized into seven English learning proficiency categories. For this study, we combined the categories EL (English Learner) and ADEL (Adult English Learner) into a single category, ``English Learners (EL) students.'' Similarly, we combined the categories EO (English Only), RFEP (Reclassified Fluent English Proficient), and IFEP (Initial Fluent English Proficient) into ``Non-English Learners (Non-EL) students.'' We excluded the remaining categories, TBD (To Be Determined) and EPU (English Proficiency Unknown), because these categories do not clearly belong to either the EL or non-EL groups. There are approximately 15\% ELLs in the entire student group.

\subsection{Experimental Design}
We conducted quantitative analysis by training and testing models under various settings to identify any potential bias. 
For the training phase, four training variants are applied for each item to train a BERT~\cite{DBLP:journals/corr/abs-1810-04805} model: only using responses from ELL students (ELL Model), only using responses from non-ELL students (Non-ELL Model), using mixed data with both ELL and non-ELL student responses which reflect the real-world distribution (Unbalanced Mixed Model), and a balanced mixed dataset with equal representation of both groups (Balanced Mixed Model). Note that, for a fair comparison, we down-sample datasets to the same size as that of ELL students so that the total number of training samples will not impact the results across these four variants. During testing, each model was tested using unbalanced mixed data, which represents the authentic composition of ELL and non-ELL students.

The number of respondents varied across test items. Based on the number of responses, we categorized the items into three groups: Group 1 has over 200,000 total responses (around 30,000 ELLs and 180,000 non-ELLs). Group 2 has about 10,000 total responses (more than 1,000 ELLs and around 10,000 non-ELLs). Group 3 has about 1,800 total responses (around 200 ELLs and 1,500 non-ELLs). Each group includes 6, 5, and 10 items, respectively.
 These groups were set up to compare the performance of BERT-base~\cite{DBLP:journals/corr/abs-1810-04805} under varying data scale conditions. Bias and disparities were calculated within each group based on the test results for all items. Finally, comparing results across the groups helps assess the impact of the data scale on AI scoring bias and disparities related to English proficiency. 

\subsection{Measure}
Building on Latif et al.'s analytical approach \cite{Latif2023}, we adopted model accuracy and Mean Score Gap (MSG) as key indicators for measuring potential bias and disparities. 
\subsubsection{Bias: Accuracy} 
Model accuracy refers to the percentage of correctly scored students’ responses out of the total testing responses, which is a fundamental and vital index for assessing model performance. In this research, we applied the Friedman test \cite{friedman1937}
 to analyze the difference in the performance of various models on mixed training data. A significant difference (\textit{p} < 0.05) indicates a difference in model quality, suggesting potential bias \cite{Latif2023} due to the varying composition of ELLs and non-ELLs in the training dataset.
\subsubsection{Disparities: Mean Score Gap} 
\begin{figure}[!ht]
    \centering
\includegraphics[width=0.7\linewidth]{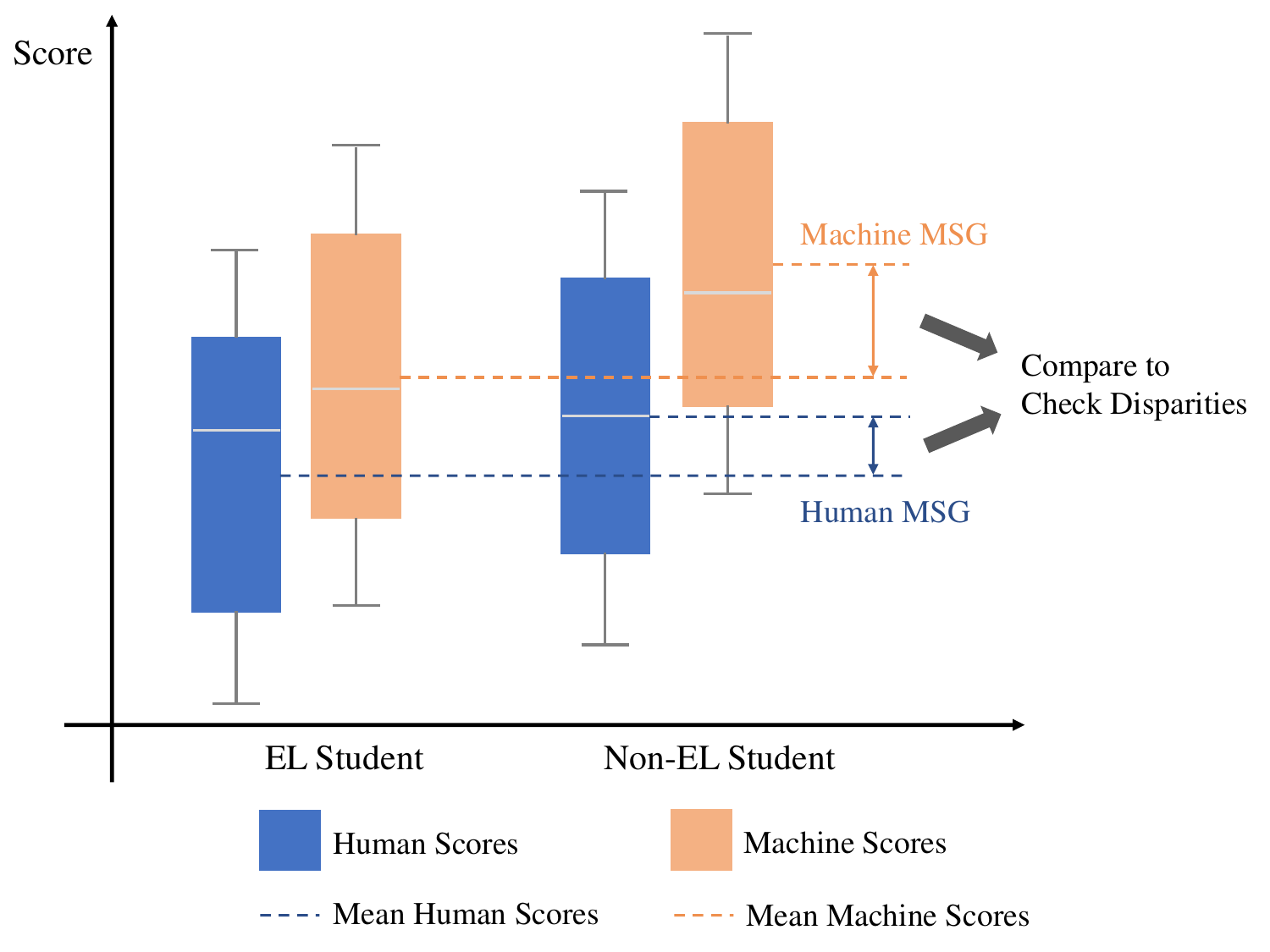}
    \caption{Analyzing disparity by comparing MSGs between human and machine scores.}
    \label{fig:msg}
    \vspace{-0.4cm}
\end{figure}

Mean score gap (MSG) refers to the gap between the average scores of two groups of students \cite{Latif2023}. In this research, the MSG in human-coded scores between the ELL and non-ELL students was taken as a baseline for evaluating the machine-generated scores. If machine-generated MSG is significantly larger or lower than the human-scored MSG, it indicates that the disparities between these ELL and non-ELL students have changed. For example, Fig.~\ref{fig:msg} shows one of the possible comparison results of MSG between human and AI scoring models, in which AI MSG is larger than human MSG, indicating that the AI scoring model widens the disparity between ELLs and non-ELLs. Due to the limited items used in the study, the Wilcoxon signed-rank test was used to detect whether there are significant differences among MSG among different AI scoring models and human scorers.

\section{Results}
\subsection{AI Bias: Accuracy Comparison}
The results of model scoring accuracy and the Friedman test are presented in Table~\ref{tab:acc}. Generally, we could observe that the models' scoring performance decreased as the dataset became smaller. Meanwhile, no significant differences emerged among models with larger ELL samples ($\textnormal{ELL}\approx30,000$: \textit{p} = 0.066; $\textnormal{ELL}\approx1,000$: \textit{p} = 0.908). However, significant differences were observed when the ELL responses were about 200 (\textit{p} = 0.006). 

Table 2 shows the post-hoc Wilcoxon signed-rank tests with Bonferroni correction (adjusted $\alpha = 0.0083$), which revealed significantly higher accuracy in the balanced mixed model compared to the ELL model (\textit{p} = 0.006, \textit{r} = 0.822). No significant differences were found among the other models.

\begin{table}[!ht]
    \centering 
    \caption{Accuracy among different models across different group sizes.} 
    \begin{tabular}{p{2.1cm} p{4cm} p{4.5cm} p{1.5cm}} 
        \hline
        \hline
        \centering{Group} & Model Type & Accuracy (\textit{Mdn (Range)}) & \textit{p}-value\\ 
        \hline
       \multirow{4}{*}{\shortstack{Group 1\\($\textnormal{ELL} \approx 30,000$)}} & \,\,Unbalanced mixed model & 0.9313 (0.8829–0.9565) & \multirow{4}{*}{0.066}\\ 
          & \,\,Balanced mixed model & 0.9294 (0.8812–0.9538) &  \\ 
          & \,\,ELL model & 0.9200 (0.8560–0.9564) &   \\ 
          & \,\,Non-ELL model & 0.9300 (0.8890–0.9592) &   \\ \hline
       \multirow{4}{*}{\shortstack{Group 2\\($\textnormal{ELL} \approx 1,000$)}} & \,\,Unbalanced mixed model & 0.8189 (0.7747–0.9715) & \multirow{4}{*}{0.908}\\ 
         & \,\,Balanced mixed model & 0.8657 (0.7568–0.9573) &  \\ 
          & \,\,ELL model & 0.8018 (0.6674–0.9614) &    \\ 
          & \,\,Non-ELL model & 0.8189 (0.7779–0.9606) & \\  \hline
           \multirow{4}{*}{\shortstack{Group 3\\($\textnormal{ELL} \approx 200$)}}  & \,\,Unbalanced mixed model & 0.7610 (0.6500–0.8722) & \multirow{4}{*}{0.006**}\\ 
         & \,\,Balanced mixed model & 0.7639 (0.6243–0.8833) &  \\ 
          & \,\,ELL model & 0.5389 (0.4167–0.8944) &   \\ 
          & \,\,Non-ELL model & 0.7425 (0.6944–0.9000) &   \\ 
        \hline
        \hline
    \end{tabular}
    \vspace{1 mm}
    \begin{minipage}{\linewidth}
        \footnotesize{\textit{Note.} * $p$ < 0.05, ** $p$ < 0.01, *** $p$ < 0.001}
    \end{minipage}
    \label{tab:acc}
\end{table}
\vspace{-10mm}

\begin{table}[!ht]
    \centering 
    \caption{Wilcoxon signed-rank test results for Group 3 ($\textnormal{ELL} \approx 200$)} 
    \begin{tabular}{
        >{\centering\arraybackslash}p{3.5cm}
        >{\centering\arraybackslash}p{2.5cm}
        >{\centering\arraybackslash}p{3cm}
        >{\centering\arraybackslash}p{3cm}}
        \hline
        \hline
        \centering{Comparison group} & \textit{Z} & Exact \textit{p}-value & Effect size \textit{r}\\ 
        \hline
       \ 1-2 & \,\,-0.059 & 1.000 & — \\
       \ 1-3 & \,\,-1.172 & 0.275 & — \\
       \ 1-4 & \,\,-0.053 & 0.992 & — \\
       \ 2-3 & \,\,-2.599 & 0.006* & 0.822 \\
       \ 2-4 & \,\,-0.561 & 0.607 & — \\
       \ 3-4 & \,\,-2.395 & 0.014 & —
        \\ 
        \hline
        \hline
    \end{tabular}
    \vspace{1 mm}
    \begin{minipage}{\linewidth}
        \footnotesize{\textit{Note.} * $p$ < 0.0083; 1= Unbalanced mixed model, 2= Balanced mixed model, 3= ELL model, 4= Non-ELL model}
    \end{minipage}
    \label{tab:acc}
\end{table}

These results indicate that when the data scale is large, no scoring bias is observed among models trained on ELL, non-ELL, unbalanced, or balanced mixed datasets. However, when the dataset is small, the accuracy of different models can differ significantly.

\subsection{AI Disparities: Mean Score Gap Difference} 
Table~\ref{tab:msg} shows MSG comparisons between AI models and human scoring across data scales. 
For large ELL samples ($\textnormal{ELL}\approx 30,000 $ and $\textnormal{ELL}\approx1,000$), no significant MSG differences emerged between AI and human scoring. With about 200 ELL responses, no MSG difference was found between the human scoring and the machine scoring for the unbalanced mixed and non-ELL models. Significant differences were observed between the human scoring and the machine scoring for the balanced mixed model (\textit{p} = 0.004) and the ELL model (\textit{p} = 0.002). Also, adding more ELLs to the training data tends to decrease the MSG. 

These findings indicate that sufficient data eliminates scoring disparities between ELL and non-ELL groups. Small datasets require training data composition matching natural distributions for optimal human-AI MSG alignment. In other words, AI scoring will not cause distorted disparities between ELLs and non-ELLs unless the composition of the training data is intentionally altered by adding more ELLs.

\begin{table} [!ht]
    \centering 
    \caption{MSG scores among different models across different group sizes and the Wilcoxon test results for the comparisons between the human scoring and each model.} 
    \begin{tabular}{p{2.2cm} p{4.2cm} p{4cm} p{1.5cm}} 
        \hline\hline
        \centering{Group} & Model Type & MSG (\textit{Mdn (Range)}) & \textit{p}-value\\ 
        \hline
       \multirow{5}{*}{\shortstack{Group 1\\($\textnormal{ELL} \approx 30,000$)}} & \,\,Human & 0.5389 (0.2883–0.7116) & —\\ 
          & \,\,Unbalanced mixed model & 0.5428 (0.2750–0.7266) & 0.438\\ 
         & \,\,Balanced mixed model & 0.5398 (0.2887–0.7221) & 0.219\\ 
          & \,\,ELL model & 0.5213 (0.2427–0.7192) & 0.563  \\ 
          & \,\,Non-ELL model & 0.5487 (0.2903–0.7202) & 0.313 \\ \hline
       \multirow{5}{*}{\shortstack{Group 2\\($\textnormal{ELL} \approx 1,000$)}} & \,\,Human & 0.3287 (0.2883–0.5786) & —\\ 
          & \,\,Unbalanced mixed model & 0.3447 (0.3175–0.6137) & 0.188\\ 
         & \,\,Balanced mixed model & 0.3112 (0.2786–0.6438) & 0.813\\ 
          & \,\,ELL model & 0.2831 (0.2574–0.5693) & 0.188  \\ 
          & \,\,Non-ELL model & 0.3372 (0.3016–0.5856) & 0.313 \\ \hline
           \multirow{5}{*}{\shortstack{Group 3\\($\textnormal{ELL} \approx 200$)}}  & \,\,Human & 0.4899 (0.2919–0.7615) & —\\ 
          & \,\,Unbalanced mixed model & 0.4663 (0.1874–0.7611) & 0.922\\ 
         & \,\,Balanced mixed model & 0.3732 (0.0305–0.5955) & 0.004**\\ 
          & \,\,ELL model & 0.1427 (0.0000–0.5976) & 0.002** \\ 
          & \,\,Non-ELL model & 0.4314 (0.1394–0.6967) & 0.131 \\ 
        \hline\hline
    \end{tabular}
    \vspace{1 mm}
    \begin{minipage}{\linewidth}
        \footnotesize{\textit{Note.} * $p$ < 0.05, ** $p$ < 0.01, *** $p$ < 0.001}
    \end{minipage}
    \label{tab:msg}
\end{table}

\section{Discussion and Conclusion}
While existing work on AI scoring bias primarily focuses on essay writing tasks, where AI often favors non-ELLs \cite{liang2023gpt,reed2023potential,wilson2024validity}, this study contributes to the field by analyzing the AI bias science constructed-response items. 
The study investigates AI scoring bias for ELLs in science assessments by comparing model accuracy and MSG across different training datasets with varying proportions of ELL and non-ELL responses. We developed four types of AI scoring models using training data that included (1) responses from ELLs, (2) responses from non-ELLs, (3) an unbalanced mixed dataset reflecting the real-world proportion of ELLs and non-ELLs, and (4) a balanced mixed dataset with equal representation of both groups, all based on BERT. These models were tested on 21 science assessment items across different data scales ($\textnormal{ELL}\approx 30,000 $, $\textnormal{ELL}\approx1,000$, $\textnormal{ELL}\approx200$). 
The results show that when training data is large enough, AI scoring models do not result in bias or distort the disparities between ELLs and non-ELLs. However, smaller datasets may introduce bias. For disparity measures, training models with naturally representative ELL/non-ELL ratios yield undistorted results.

Our findings are unsurprising and contribute to the ease of unnecessary fears of AI bias, which are critical for all populations to benefit from AI\cite{2024Zhai}. Although many studies have found automatic scoring bias for ELLs, similar to this study, Wilson and Huang (2024) found that when using an automatic scoring system for writing tests, the system did not exhibit unique bias against ELLs but rather replicated biases present in human scoring \cite{wilson2024validity}. Reed and Mercer (2022) also found no evidence of bias for ELLs when comparing interim assessments scored by teachers or expert raters to predict student performance on a summative writing test \cite{reed2023potential}.
Meanwhile, 
empirical studies that found English proficiency to be an important factor that impacts the automatic scoring or categorization results of student writing tasks, are primarily focused on essay writing tasks \cite{liang2023gpt}  \cite{reed2023potential} \cite{wilson2024validity}. 
These tasks asked students to write quite long responses, and their writing proficiency is often a key evaluation criterion, which is highly related to their English proficiency. Unlike essay writing assessments, science constructed-response items are typically shorter and less dependent on language proficiency, reducing the impact of language proficiency on AI scoring. 
Moreover, science assessments may include multimodal components, such as diagrams, tables, and symbols, which support students' understanding of the concepts in the question and help the accurate interpretations of the students' understanding beyond their language proficiency \cite{grapin2020} \cite{grapin2022multimodal}.Thus, the automatic scoring of these responses is less likely to be influenced by students' English proficiency compared to previous research on essay writing tasks. Our findings therefore suggest that AI scoring neither amplifies existing bias nor distorts true disparities for ELLs.  

Besides, the results could be the potential consequence of the fact that machines imitate humans, as Nesbitt and Quesen also argued \cite{nesbitt2023}, since we found that natural ratio of ELL/non-ELL lead to undistorted disparities in smaller datasets. According to the results, it seems that the unbalanced mixed model and non-ELL model replicate the scoring pattern observed in the human/training data, whereas incorporating ELL samples into the training data helps reduce the scoring gap between the ELL and non-ELL groups. However, at this stage, it is difficult to determine whether the decrease in MSG indicates less bias toward ELLs or if it is actually due to other potential issues, such as lowering the scores for non-ELLs. While this should be further explored in the future, until more empirical evidence emerges, we recommend using adequately sized datasets that reflect real-world ELL proportions as training data to minimize additional bias in science assessment.

\section{Limitation and Future Study}
Although the results of this study show that AI scoring generally does not lead to biased results for ELLs and non-ELLs in science constructed response assessments, it is important to note that AI bias does not solely stem from the factor investigated (i.e., the training dataset in this study). Schwartz et al. \cite{schwartz2022towards} classify AI biases into three groups: systemic, statistical, computational, and human biases. Systemic biases, such as sexism and racism, are inherited in datasets. They arise from the historical and institutional norms, as well as societal practices that privilege certain groups of people. Statistical and computational biases are caused by the lack of representation or misrepresentation of certain groups during AI training. Human biases, on the other hand, are mostly related to the user and AI interaction, such as biases in prompts, interpretation, confirmation, and the use of the outputs.
Therefore, future studies should continue to address the potential AI bias for ELL students, considering all these possible dimensions. Moreover, future research can provide an even more nuanced approach by exploring whether AI scoring bias and disparities exist for different levels of English proficiency, such as WIDA's proficiency level distinctions, rather than a broad ELL/non-ELL binary \cite{wida2020wida}.

Finally, while it is crucial to explore the potential bias in automated scoring, it is even more important to step back and reconsider the possible embedded biases in the entire assessment process, from creating items to rubric design and the underlying human assumptions about race, ethnicity, and language of those who participate in the assessment \cite{nesbitt2023}. Future research is needed to provide a more comprehensive perspective on more inclusive science assessments that accurately capture all students' understanding of science concepts. 

\section*{Acknowledgment}
The research reported here was supported by the Institute of Education Sciences, U.S. Department of Education, through Grants R305C240010 (PI Zhai) and R305A240356 (PI Liu). The opinions expressed are those of the authors and do not represent the views of the Institute or the U.S. Department of Education. Additionally, we acknowledge the use of OpenAI's ChatGPT for polishing the language and enhancing the clarity of this manuscript.

\bibliographystyle{splncs04}
\bibliography{reference.bib}
%




\end{document}